\documentclass[letterpaper, 10 pt, conference]{ieeeconf}  

\IEEEoverridecommandlockouts                              

\usepackage{graphicx}
\usepackage{subcaption}

\usepackage{amsmath} 

\title{\LARGE \bf
Toward Efficient Visual Gyroscopes: Spherical Moments, Harmonics Filtering, and Masking Techniques for Spherical Camera Applications
}

\author{Yao Du$^{1}$, Carlos M. Mateo$^{2}$, Mirjana Maras$^{1}$, Tsun-Hsuan Wang$^{3}$ , Marc Blanchon$^{1}$, Alexander Amini$^{3}$,\\ Daniela Rus$^{3}$, Omar Tahri$^{2}$  
\thanks{This work was partially supported by Capgemini Engineering.}
\thanks{$^{1}$Hybrid Intelligence part of Capgemini Engineering
        {\tt\small \it{\{first\_name.last\_name\}}@capgemini.com}}%
        \thanks{$^{2}$Universit\'e de Bourgogne, CNRS UMR 6303 ICB, Dijon, France
        {\tt\small \it{\{carlos-manuel.mateo-agullo,omar.tahri\}}@u-bourgogne.fr}}%
\thanks{$^{3}$Computer Science and Artiﬁcial Intelligence Lab, Massachusetts Institute of Technology
        {\tt\small \it{\{{tsunw,amini,rus}\}}@mit.edu}}%
}

\begin{document}

\maketitle
\thispagestyle{empty}
\pagestyle{empty}

\begin{abstract}

Unlike a traditional gyroscope, a visual gyroscope estimates camera rotation through images.
The integration of omnidirectional cameras, offering a larger field of view compared to traditional RGB cameras, has proven to yield more accurate and robust results. However, challenges arise in situations that lack features, have substantial noise causing significant errors, and where certain features in the images lack sufficient strength, leading to less precise prediction results.

Here, we address these challenges by introducing a novel visual gyroscope, which combines an Efficient Multi-Mask-Filter Rotation Estimator(EMMFRE) and a Learning based optimization(LbTO) to provide a more efficient and accurate rotation estimation from spherical images. 
Experimental results demonstrate superior performance of the proposed approach in terms of accuracy. The paper emphasizes the advantages of integrating machine learning to optimize analytical solutions, discusses limitations, and suggests directions for future research.
\end{abstract}

\section{INTRODUCTION}
A classical problem in robotics is the estimation of the orientation of a camera. 
By analyzing the features of two or more consecutive images, a Visual Gyroscope (VG) can determine the orientation and angles of the camera, instead of relying on mechanical structures like traditional gyroscopes. 
With its ability to provide precise measurements and real-time tracking ~\cite{caron2018spherical,corke2004omnidirectional}, the visual gyroscope is a powerful tool for a wide range of industries and fields, from stabilizing cameras and drones to navigating autonomous vehicles and spacecraft.

Generally, visual gyroscopes can be categorised by different types of sensors, estimation methods, and feature extraction techniques. 
Employed sensors include monocular cameras~\cite{hartmann2015visual}, stereo cameras~\cite{oskiper2007visual}, panoramic cameras~\cite{andre2022photometric},  RGB-D cameras~\cite{ruotsalainen2013two}, or combinations thereof~\cite{chen2015single}. 
Omnidirectional cameras have emerged as a valuable tool for VG, as they can capture a full 360-degree view of the surroundings, resulting in offering richer information.

Moreover, based on different methods, VGs can be classified into types based on extended Kalman filters (EKF) ~\cite{kyrki2005integration,kragic2006initialization}, sequential Monte Carlo methods, particle filters (PF)~\cite{ababsa2007robust,schon2005marginalized,sadeghzadeh2014particle}, optical flow~\cite{li2021gyroflow,hong2008visual,goppert2017invariant}, feature-based~\cite{miao2021univio,yu2022tightly}, or even Fourier transform based methods~\cite{birem2018visual,schairer2009increased}.
However, most of the methods above include non-linear state equations and non-Gaussian noise assumptions, which impact the resulting accuracy and efficiency. 
In addition, these approaches demand greater computational and memory resources, while also being sensitive to minor changes in dynamic environments.

\begin{figure}
\centering
\includegraphics[width=0.7\columnwidth]{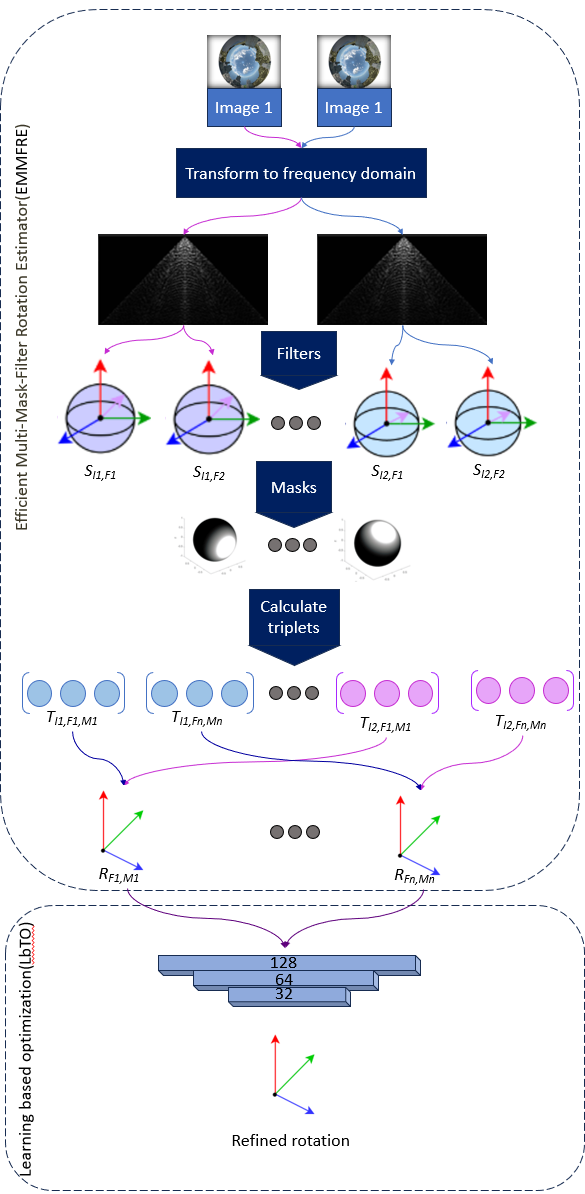}
\caption{{\bf Overview of the novel efficient visual gyroscope.} The proposed visual gyroscope method consists of two key computation blocks: Efficient Multi-Mask-Filter Rotation Estimator (EMMFRE) and the Learning based optimization (LbTO), which conjunctively lead to a more accurate and efficient final image rotation estimation.}
\label{fig:Method Overview}
\end{figure}

To address these challenges, this paper proposes a novel method, as illustrated in Fig. \ref{fig:Method Overview}, the Fast Visual Gyroscope (FVG), which computes more accurate and efficient 3D orientations of the camera for a given image with respect to a reference. 
Our approach offers a faster and more accurate computation of rotation estimates thanks to the efficiency gain from the new analytical step and the accuracy gain from the learning-based optimization of rotation estimates. The efficacy of our method was demonstrated against baseline visual gyroscopes method~\cite{andre2022photometric,chen2021wide}, highlighting the advantages of the fast visual gyroscope.

\section{Related Works}\label{RelatedWork}

According to different operating principles, visual gyroscopes can be divided into different categories: the ones
based on Extended Kalman filter(EKF)~\cite{kyrki2005integration,kragic2006initialization}, methods based on optical flow~\cite{li2021gyroflow,goppert2017invariant}, using image features~\cite{miao2021univio,yu2022tightly}, Fourier transform~\cite{birem2018visual,schairer2009increased}, particle filtering~\cite{ababsa2007robust,ababsa2006robust} and hybrid solutions~\cite{didier2008hybrid,camposeco2018hybrid}. 

Kyrki et al.~\cite{kyrki2005integration} integrated model-based and model-free cues using EKF but faced robustness issues with outliers, while Kragic et al.~\cite{kragic2006initialization} extended their work by developing a method for automatic initialization of pose tracking based on robust feature matching.
However,the approximation in EKF can lead to poor representations of the nonlinear functions. 
The sequential Monte Carlo method, or particle filters~\cite{ababsa2007robust} could provide improved robustness over the Kalman filters. 
Qian et al.~\cite{qian2002bayesian} describe an ad-hoc method for incorporating gyroscope measurements.
And Sadghzadeh et al.~\cite{sadeghzadeh2014particle} employ the Bayesian based PF approach to estimate inertia tensor due to nonlinear and non-Gaussian models. 
On the other hand, optical flow represents the distribution of apparent velocities of brightness patterns in an image~\cite{li2021gyroflow,goppert2017invariant,zhou2023event}, and is used to estimate the projected motion of the relative displacement between the camera and the objects. 
Comparing to which, feature-based gyroscopes~\cite{tarrio2015realtime,miao2021univio,hua2019feature,yu2022tightly} use the detection and tracking of distinctive image features, such as corners, edges, or blobs, to estimate rotation, 
They can achieve higher accuracy and robustness than optical flow gyroscopes, but require more computation and memory resources. 
However, optical flow or feature-based methods are easily affected by small changes in the dynamic environment, as well as when the images change significantly (such as in the case of large movements)~\cite{schairer2009increased}. 
Based on harmonic analysis~\cite{chirikjian2000engineering,miller2001group}, Makadia et al.~\cite{makadia2003direct} proposes a framework for studying image deformation applicable in the plane and on the sphere. These deformations have also been explored in learning-based methods~\cite{fernandez2020corners,wu2020depth,wu2022depth} and have achieved great results.
Similarly, Burel et al.~\cite{burel1995determination} determine the 3D orientation from normalizing tensors which are obtained from spherical harmonics coefficients. Chirikjian et al.~\cite{chirikjian2000engineering} shows good applications using this method. 


In summary, the application of visual gyroscopes is still confronted with several limitations. These include a heavy reliance on visual features, which can pose challenges in feature-deficient or low-light conditions. Furthermore, the accuracy of visual gyroscopes is susceptible to variations in camera pose and the geometric relationship between the camera and its surroundings. Lastly, the computationally intensive nature of visual feature tracking and analysis limits the applicability of this technology to low-power devices.

A novel approach utilizing spherical moments has recently been developed, which constructs a feature called "triplets" to estimate camera rotation. This method demonstrates robustness in scenarios with sparse features. Additionally, it offers the advantage of rapid computation. Essentially, triplets can be conceptualized as sets of three-dimensional points in a Cartesian coordinate system. By computing triplets from two images, the rotation matrix between the point clouds can be calculated using methods such as Procrustes analysis, which corresponds to the camera's rotation between the two image captures. Further details regarding this methodology can be found in ~\cite{9305374}.

\section{Contribution}\label{Contribution}

The proposed Fast Visual Gyroscope (FVG) method hinges on three key innovations. Firstly, a novel feature extraction pipeline is introduced, leveraging spherical harmonics coefficients for robust and discriminative feature representation. By applying frequency domain filtering and a multi-mask strategy, we enhance feature invariance and global representation. Secondly, a hybrid approach combining analytical and learning-based components is employed. Raw rotation estimates from an analytical solution are fed as input to a Multi-Layer Perceptron (MLP), enabling efficient and accurate rotation estimation. Thirdly, an adaptive feature selection mechanism based on the MLP is implemented to optimize feature combination for different input conditions. Through extensive data augmentation and training on a simulated dataset, the MLP is trained to effectively learn the complex mapping between input features and ground truth rotations.
Finally, the proposed pipeline is tested on experiments, providing a comprehensive evaluation of its efficacy.  

\section{Methods}\label{methods}

According to the pipeline (Fig.~\ref{fig:Method Overview}) proposed in this paper, the initial step involves transforming the spherical image into the spherical harmonics domain, followed by filtering operations within this domain. Subsequently, spherical moments are directly computed in the spherical harmonics domain. The calculated spherical moments are then combined linearly to obtain masked spherical moments. Next, a rotation estimate is derived from the masked spherical moments, and this estimation is utilized as input for optimizing a multilayer perceptron.

Here we describe in more detail the three crucial steps of our algorithm. The first part introduces a rapid method for directly computing spherical moments in the spherical harmonics domain. The second part outlines the approach of obtaining masked spherical moments through linear combinations. The third part explores the structure of the MLP, loss functions, and training strategies.

\subsection{From Spherical Harmonics to Spherical Moments}

\begin{figure}[!htb]
\centering
\includegraphics[width=0.99\linewidth]{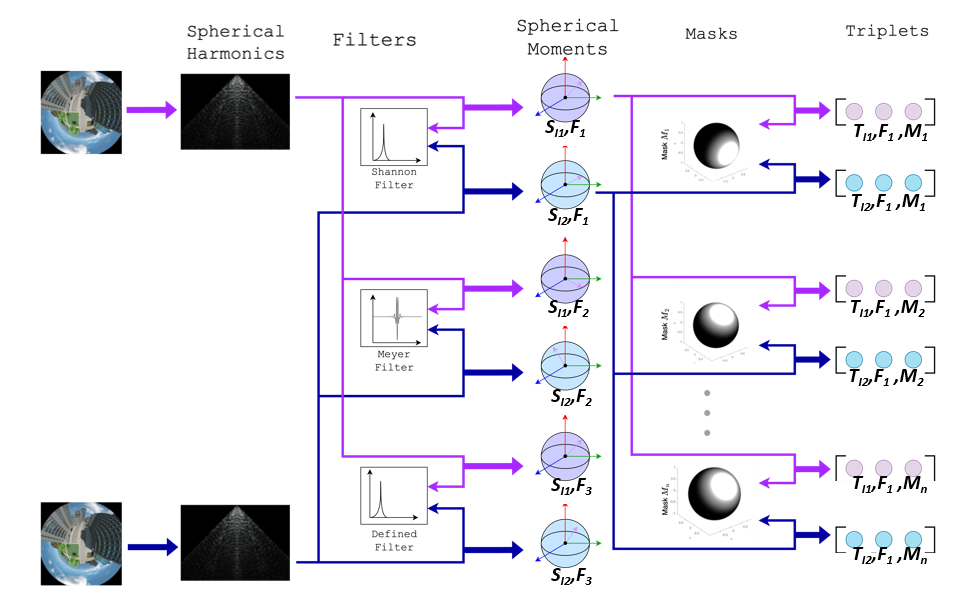}
\caption{{\bf Image pre-processing for the analytical computation of triplets derived from spherical moments.} First filtering and masking (total of 100 masks) are applied to the images, and then the spherical moments are computed before the feature triplets.}
\label{fig:Masks and filters}
\end{figure}

Generally, the method to compute spherical moments is composed of two steps: firstly, the image in the frequency domain is transformed to the spatial domain, via the inverse Fourier transform. Secondly, the image is projected onto a sphere, and then the spherical moments of the image are calculated. \par
Originally, spherical moments in image domain is defined as follows:
\begin{equation}
m_{i j k}=\iint_{s} x_{s}^{i} y_{s}^{j} z_{s}^{k}I(s) ds.
\label{mijk}
\end{equation} 

\noindent
Thus, moments are computed as the integral of the image over the surface of an unitary sphere.
But if the spherical moment could be directly calculated from the spherical harmonic coefficients, the calculation speed would be accelerated. The definition of spherical harmonics coefficients $\hat{I}_{l m}$:

\begin{equation}
\label{SHcoefficient2Image}
\hat{I}_{l m}=\iint_{s} I(s) Y_{l m}^{*}(s) ds,
\end{equation}
where $Y_{l m}^{*}$ is the conjugate of spherical harmonics of degree $l$ and order $m$ (integer between $-l$ and $l$),  $I$ is original image over the unity sphere surface $s=(\theta,\varphi)$.
\par

The calculation of $Y_{lm}$ is given by the following formula:

\begin{small}
\begin{equation}
Y_{lm}(\theta,\phi) = (-1)^m \sqrt{\frac{2l+1}{4\pi} \frac{(l-m)!}{(l+m)!}} P_l^m(\cos\theta) e^{im\phi} = \Psi,
\label{ylm}
\end{equation}
\end{small}

\noindent where $\theta$ is the elevation angle, $\phi$ is the azimuthal angle,  $i$ is the imaginary unit, and $P_l^m(\cos\theta)$ is the associated Legendre polynomial, which can be calculated using the recursion formula:

\begin{small}
\begin{equation}
\begin{split}
& (l-m+1)P_{l+1}^m(\cos\theta) - (l+m)P_{l}^m(\cos\theta) + \\
& (l+m) \frac{(l-m+1)}{\sin\theta} P_l^{m-1}(\cos\theta) = 0,
\end{split}
\end{equation}
\end{small}

\noindent with the following initial conditions,

\begin{equation}  
\begin{cases}
 P_l^l(\cos\theta) = (-1)^l (2l-1)!! \gamma\\
 P_l^{-l}(\cos\theta) = (-1)^l \frac{(2l-1)!!}{(2l)!!} \gamma^{m/2} \frac{d^{|m|}}{d(\cos\theta)^{|m|}} (-\gamma)^l
\end{cases},
\end{equation}

\noindent where $\gamma = (1-\cos^2\theta)$. The normalization factor $\sqrt{\frac{2l+1}{4\pi} \frac{(l-m)!}{(l+m)!}}$ in equation~(\ref{ylm}) ensures that the spherical harmonics are orthonormal.

Substituting the solution for $I(s)$ from equation~(\ref{SHcoefficient2Image}) into equation~(\ref{mijk}), the spherical moments can be computed from spherical harmonics coefficients as:
\begin{equation}
\label{SHcoefficient2moment}
\begin{aligned}
m_{i j k} &=\sum_{l} \sum_{m} \iint_{s} x^{i} y^{j} z^{k} \hat{I}_{l m} \Psi d s, \\
&=\sum_{l} \sum_{m} \hat{I}_{l m} C^{ijk}_{l m},
\end{aligned}
\end{equation}

\noindent where we define the moments coefficient $C_{l m}^{i j k}$ as:

\begin{equation}
\begin{array}{ll}
 \begin{aligned}
C_{l m}^{i j k} &= \iint_{s} x^{i} y^{j} z^{k} \Psi d s, \\
&=\iint_{s} (\sin\theta)^{i+j+1} \cos \varphi^{i} \sin \varphi^{j} \cos \theta^{k} \Psi d \theta d \varphi .
\end{aligned}
\end{array}
\end{equation}

\noindent where, as a recall, $\Psi = Y_{l m}(\theta, \varphi)$. 

Equation (\ref{SHcoefficient2moment}) provides our analytical closed-form expression for the set of spherical moment coefficients of different orders. This new analytical expression transforms convolution and integration operations into a multiplication operation, thus greatly reducing the computational complexity of the final expression. Furthermore, since the function $C_{l m}^{i j k}$ that we introduce is a basis function, we can reduce computational complexity by storing the values for each needed $(l,m,i,j,k)$ tuple.

\subsection{Fast Implementation of Mask on Spherical Moments}

A known common issue when using visual gyroscopes for the estimation of rotation on global features is that the presence of non-overlapping regions in two images could reduce the accuracy in ego-motion visual estimation. Since our analytical expression (\ref{SHcoefficient2moment}) would suffer from this problem as well, we propose the use of different masks to reduce the influence of non-overlapping regions before calculating the triplets.

\begin{figure}[t]
	\centering
	\begin{subfigure}{0.3\linewidth}   
		\centering 
		\includegraphics[width=\linewidth]{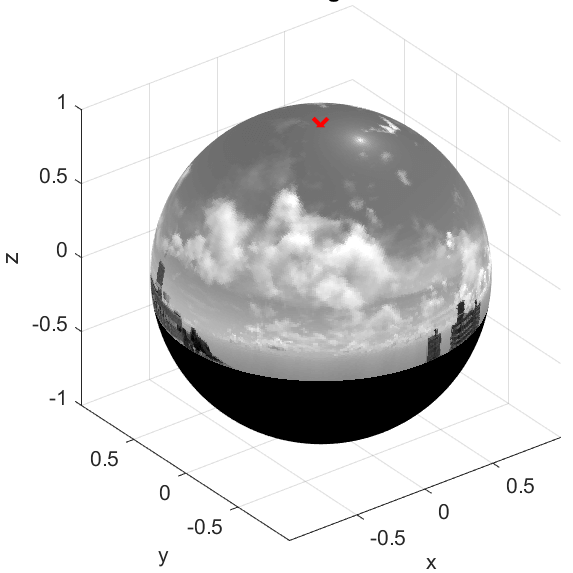}  \caption{}
        \label{IcosahedralSample}
	\end{subfigure} 
 	\begin{subfigure}{.3\linewidth}   
		\centering
		\includegraphics[width=\linewidth]{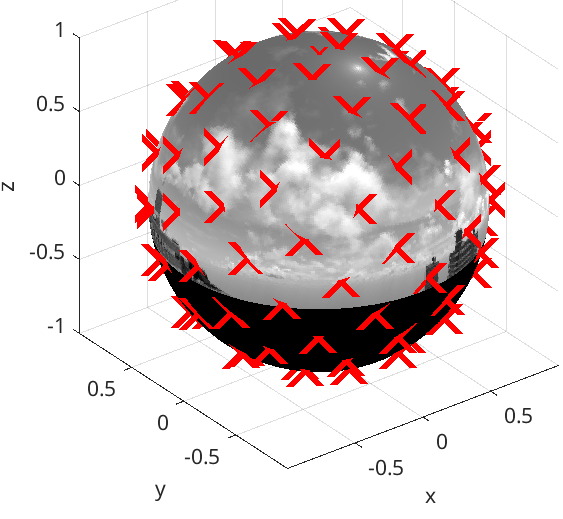}        
        \caption{}
        \label{IcosahedralSampleSphereWhole}
	\end{subfigure} 
	\begin{subfigure}{.3\linewidth}   
		\centering
		\includegraphics[width=\linewidth]{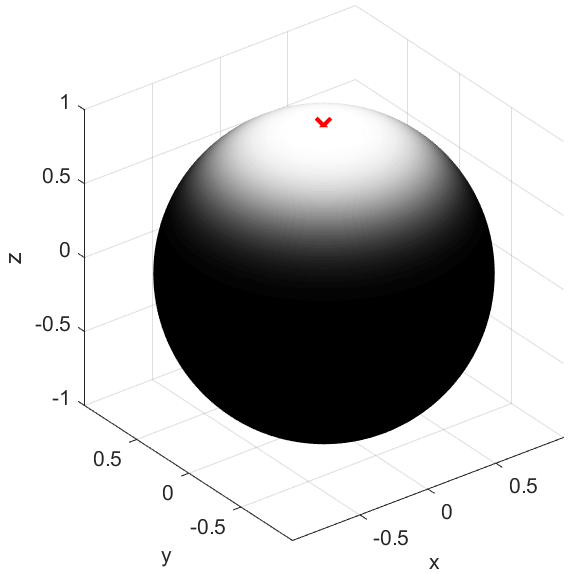}      
        \caption{}
        \label{MaskOnSphereImage}
	\end{subfigure}
    \begin{subfigure}{.3\linewidth}   
		\centering 
		\includegraphics[width=\linewidth]{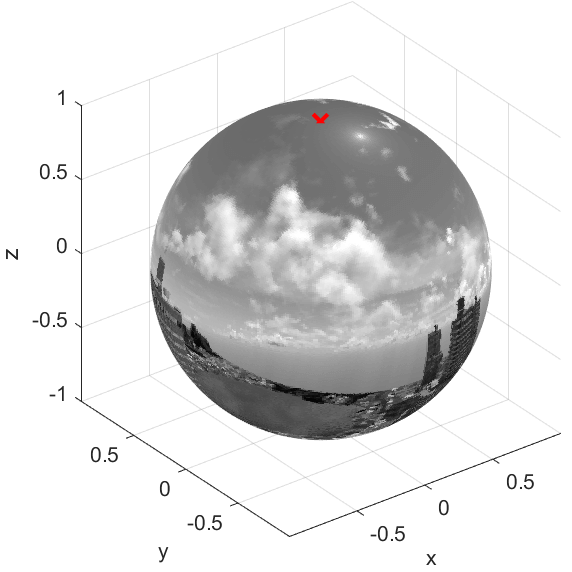}
        \caption{}
        \label{SphereMask}
\end{subfigure}
	\begin{subfigure}{.3\linewidth}   
		\centering 
		\includegraphics[width=\linewidth]{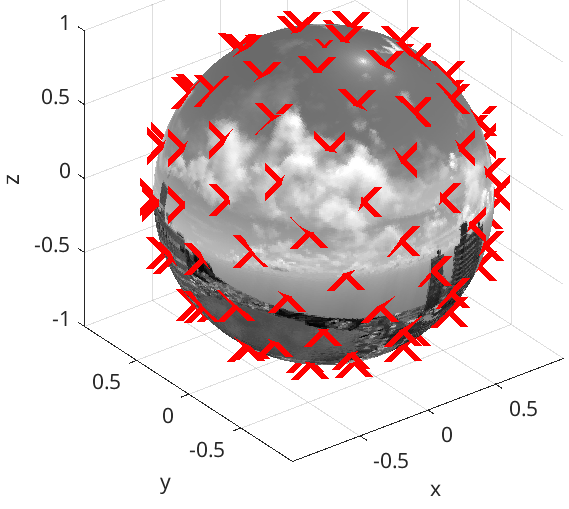}
        \caption{}
        \label{SphereImage}
	\end{subfigure}
    \begin{subfigure}{.3\linewidth}   
		\centering 
		\includegraphics[width=\linewidth]{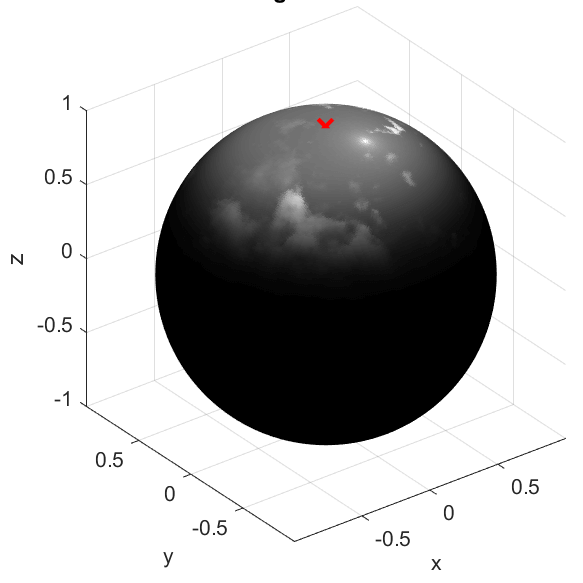}
        \caption{}
        \label{SphereImagewhole}
	\end{subfigure}
    \caption{{\bf Masking method explained.} (a) and (d) show the images on a half sphere and around the complete spherical surface. (b) presents an icosahedral sample delicately positioned on a half sphere. The figure (e) extends its scope to showcase the icosahedral sample enveloping the entire surface of a sphere. (c) and (f) introduce a mask on a sphere and an image overlaid with a mask.}
	\label{fig:3masks} 
\end{figure}

A sampling method uniformly distributed according to azimuth and elevation will result in too many sampling points close to the poles and too few sampling points near the equator.
Therefore, we used the sampling method of icosahedral distribution to avoid this problem as shown in Fig.~\ref{fig:3masks}.

\begin{figure}[b]
	\begin{subfigure}{.3\linewidth}   
		\centering 
		\includegraphics[width=\linewidth]{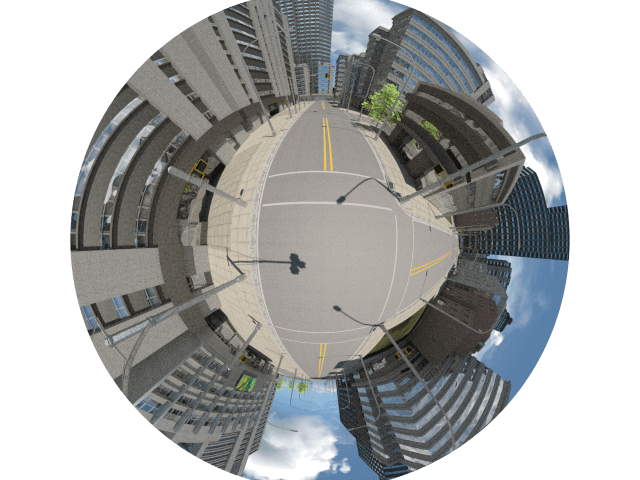}
        \caption{}
	\end{subfigure}
	\begin{subfigure}{.3\linewidth}   
		\centering 
		\includegraphics[width=\linewidth]{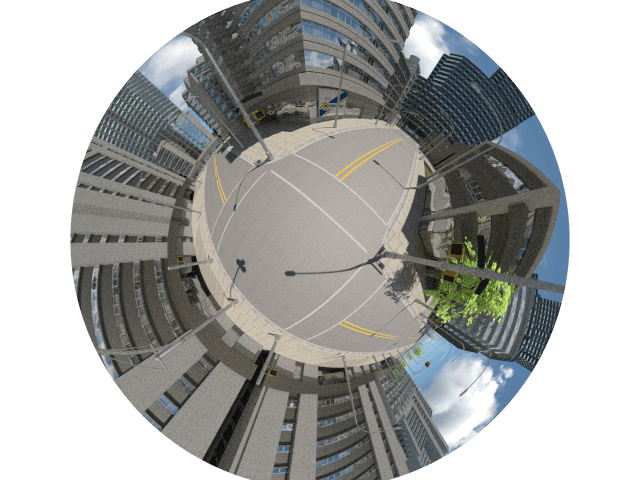}
        \caption{}
	\end{subfigure}
    \begin{subfigure}{.3\linewidth}   
        \centering 
        \includegraphics[width=\linewidth]{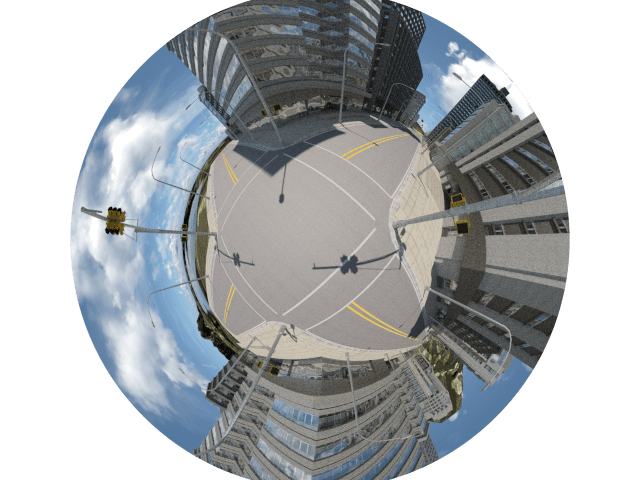}
        \caption{}
	\end{subfigure}
    \caption{{\bf Images generated from the Blender simulation environment.} We generate 500 images for the training and 150 for the testing of the MLP, by randomly sampling the rotation metrics. (a)-(c) show three examples of the generated images.}
	\label{fig:imagesFromBlender} 
\end{figure}

Taking into account the targeted application, a well adapted mask is the one of a round shape, with the weight 1 around the region center, and decreasing softly around the region border.
Inspired by~\cite{bakthavatchalam2018direct}, a good candidate that holds these conditions is defined as:
\begin{equation}
\label{maskideal3d}
 \begin{cases} W(z_s) = 0 & \left(-1 <z_s<z_0-r\right) \\ W(z_s) =e^{-\frac{\left(z-\left(z_0+r\right)\right)^2}{\sigma^2}} & \left(z_0-r \leq z_s \leq z_0+r\right) \\ W(z_s) = 1 & \left(z_0+r<z_s<1\right) \\  \end{cases}.
\end{equation}
The mask's shape is determined by parameters $\sigma$ and range $r$, while $z_0$ represents the center of the mask.
Such a shape is well adapted to recover the rotation around the z-axis.
However, its formulation is quite complex for computing moments corresponding to the selected region.
Instead of using equation~(\ref{maskideal3d}), masks under polynomial form on the variate $z_s$ can be used:
\begin{equation}
W(x, y, z) = \sum_{l=0}^n \sum_{m=0}^n \sum_{p=0}^n a_{lmp} \cdot x^l \cdot y^m \cdot z^p,
\label{mskPoli3d}
\end{equation}

\noindent where $a_{lmp}$ are the coefficient of the polynomial.
They are defined such that the equation~(\ref{mskPoli3d}) approximates the shape of the mask defined by the equation~(\ref{maskideal3d}).
According to the equation~(\ref{mijk}) and the equation~(\ref{SHcoefficient2Image}):
\begin{equation}
\begin{aligned}
m_{i j k}^{mask} &=\sum_{l} \sum_{m} \iint_{s} x^{i} y^{j} z^{k} \hat{I}_{l m} e^{-\left(\frac{z-z_{0}}{\sigma}\right)^{g}} \Psi d s, \\
&=\sum_{l} \sum_{m} \hat{I}_{l m} \iint_{s} x^{i} y^{j} z^{k} \sum_{h=0}^{\infty} \frac{\left(\frac{z-z_{0}}{\sigma}\right)^{g h}}{h !} \Psi d s, \\
&=\sum_{l}  \sum_{m} \hat{I}_{l m} \Upsilon_{mask}^{i j k},
\end{aligned}
\end{equation}

\noindent where the coefficient defined as:
\begin{equation}
\begin{aligned}
\Upsilon_{mask}^{i j k} 
&= \iint_{s} x^{i} y^{j} z^{k} \Psi \sum_{h=0}^{\infty} \frac{\left(\frac{z-z_{0}}{\sigma}\right)^{g h}}{h !} d s .
\end{aligned}
\end{equation}

\noindent For the last sum part and the item with $z^{k}$:
\begin{equation}
z^{k} \sum_{h=0}^{\infty} \frac{\left(\frac{z-z_{0}}{\sigma}\right)^{g h}}{h !},
\end{equation}
\noindent can be written as a linear combination of $z^{p}$, namely:
\begin{equation}
z^{k} \sum_{h=0}^{\infty} \frac{\left(\frac{z-z_{0}}{\sigma}\right)^{g h}}{h !}
= \sum_{p=k}^{k+q} a_p z^{p}.
\end{equation}
\noindent Therefore, we can get the coefficient with a mask from a linear combination of the higher order coefficients without the mask:
\begin{equation}
\begin{aligned}
\Upsilon_{mask}^{i j k} &= \iint_{s} x^{i} y^{j}  \Psi \sum_{p=k}^{k+q} a_i z^{p} d s , \\
&=\sum_{p=k}^{k+q} a_i \iint_{s} x^{i} y^{j}  z^{p} \Psi  ds ,\\
&= \sum_{p=k}^{k+q} C^{i j k}_{lm} .
\end{aligned}
\end{equation}


Using the coefficient $\Upsilon_{mask}^{ijk}$ of spherical moments, it is possible to directly obtain masked spherical moments and subsequently calculate a set of triplet features. These triplet features provide valuable information about the relative positions and orientations of objects, which can aid in rotation estimation.
With the set of triplet features, it is possible to obtain an analytical solution for rotation estimation. This solution can be further enhanced using various techniques, such as regularization or noise modeling, to improve its accuracy and robustness.

Overall, by leveraging the coefficient $\Upsilon_{mask}^{ijk}$ of spherical moments, one can obtain both spherical moments and triplet features, which can be used to obtain a reliable and accurate analytical solution for rotation estimation. 

\begin{figure*}[!htb]
\centering
\begin{subfigure}{.49\textwidth}
    \centering
    \includegraphics[width=0.99\linewidth]{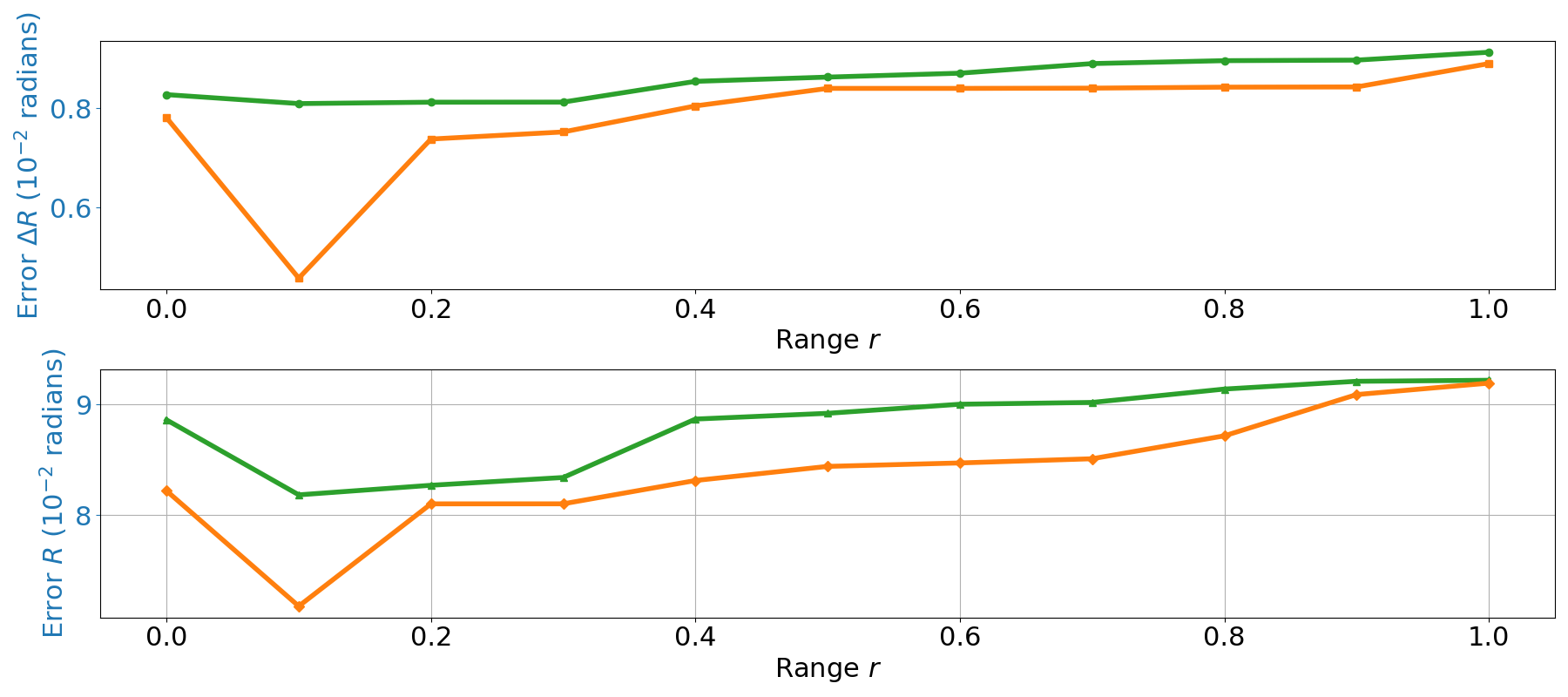}
    \caption{}
    \label{fig:compare_seqPure3D_half}
\end{subfigure}%
\begin{subfigure}{.49\textwidth}
    \centering
    \includegraphics[width=0.99\linewidth]{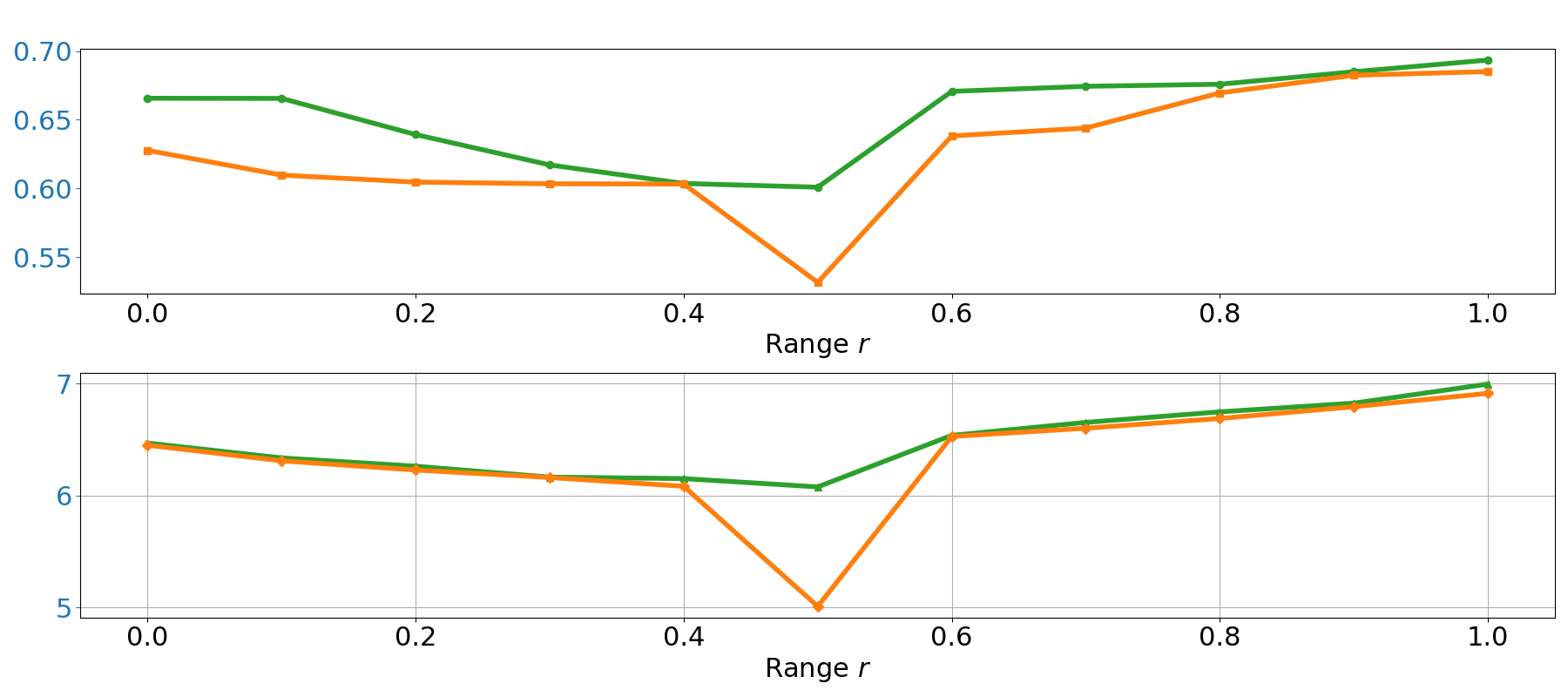}
    \caption{}
    \label{fig:compare_seqPure3D_whole}
\end{subfigure}%
\caption{{\bf Analysis of the impact of mask range on the accuracy of rotation estimations.} The full analytical-LbTO method (fast visual gyroscope) is compared to the analytical method alone, to reveal optimal range value and significantly higher accuracy of our proposed analytical-LbTO method. (a) displays the comparison of error for various values of the range $r$ in a pure rotation sequence with half sphere images, (b) shows comparison of error for various values of the range $r$ in pure rotation sequence with whole sphere image.}
\label{fig:OtherResQuant}
\end{figure*}

\subsection{LbTO: Learning-based Triplet Optimizer}

To further increase the accuracy of the predicted rotation estimation on spherical images, we introduce the third and last step in our method: a neural network based optimization of the type and number of masks and filters. Specifically, we train an MLP to choose the masks and filters that minimize the error between the predicted and ground truth rotation vectors. The MLP is trained and tested with synthesized fisheye camera data from Blender as shown in Fig.~\ref{fig:imagesFromBlender}.

In terms of the neural network architecture, we use a three-layer 128x64x32 MLP (excluding input and output layers). During the training process, we combine a decaying learning rate, SWA learning rate schedules, and the Adam optimizer in order to  accelerate the MLP learning convergence, prevent overfitting, and enhance the generalization performance. And for the loss function we use Mean Squared Error (MSE).

Addressing the challenge of discontinuity in MSE loss functions requires careful consideration of degrees of freedom (DoF). While augmenting DoF presents a potential solution, it necessitates careful balance due to its impact on training time and real-time performance. For instance, Zhou et al.~\cite{zhou2019continuity} proposed 5D and 6D continuous representations to mitigate discontinuity issues in computer vision deep learning approaches, focusing exclusively on rotational aspects. However, considering our emphasis on estimating rotations between adjacent images, where estimates typically hover near zero, we opt for the axis-angle representation. This choice ensures practicality without compromising accuracy, aligning seamlessly with the objectives of our study.

\section{Experiments}\label{experiments}

Due to the potential for errors in the ground truth caused by temperature drift and zero drift in the inertial measurement unit in real-world environments, all experiments are conducted in the Blender simulation environment.

The experiment involves pure 3D rotation motion, which tests the algorithm's ability to estimate rotation accurately without the influence of other factors. 

We perform the experiment with a dataset generated in the Blender simulation environment, 
which consisted of 500 images, 30\% of which are used for testing and 70\% for training.
The rapidity of our proposed fast visual gyroscope method is demonstrated by its implementation with 100 masks, taking only 20 milliseconds to apply all masks.

\subsection{Investigate the dimensions of the mask}
To study the impact of the learning based optimization step in our three-step visual gyroscope approach, we conduct an experiment in which we evaluate the impact of the range on the accuracy of predicted rotation estimates. Our experiment demonstrates the criticality of the learning based optimization, which leads to a much more accurate rotation estimation than had we used the analytical steps alone. Fig.~\ref{fig:compare_seqPure3D_half} shows that when only half of the spherical image is available, we can determine that $r = 0.1$ is the optimal choice for minimizing errors through empirical experimentation with different range values. In this case, selecting a mask with a smaller radius is beneficial to prevent interference from the edges of the spherical image, which can lead to errors. This approach is effective for estimating both the rotation difference, $\Delta R$, and the overall rotation, $R$, between two images.

The rationale behind this finding is that when considering only half of the sphere, the value $r = 0.1$ ensures that the mask does not extend to the boundaries of the spherical image. 
If we were to increase the value of $r$ further, the impact of the spherical image boundaries on error would become more significant relative to the reduction in error achieved by expanding the field of view. 
Therefore, the choice of $r = 0.1$ strikes a balance between these two effects, ultimately minimizing the error.


Similarly, the data in Fig.~\ref{fig:compare_seqPure3D_whole} clearly indicates that when considering the entire spherical surface, through experimentation with different range values, we can ascertain that $r = 0.5$ emerges as the most effective choice for minimizing errors. 
This conclusion holds true not only for the estimation of rotation $\Delta R$ but also for the cumulative rotation estimation $R$ between two images.

The rationale behind this outcome remains consistent with the prior scenario. 
When dealing with the entire spherical surface, the value $r = 0.5$ notably amplifies the reduction in error achieved through the expansion of the field of view. 
This reaffirms the earlier theory that $r = 0.1$ minimizes the error on the half spherical surface.
It is worth noting that the analytical solution method can suffer from inaccuracies due to its dependence on the used masks. 
However, the results obtained using the LbTO are closer to the ground truth, indicating the effectiveness of the proposed approach in estimating the rotation angles.
This conclusion is crucial for our problem, as it helps determine the optimal parameter configuration to achieve as accurate a rotation estimation as possible. Such analysis contributes to improving the performance and accuracy of image processing algorithms.

\begin{figure*}[htb]
\centering
\begin{subfigure}{.33\textwidth}
  \centering
    \includegraphics[width=0.99\linewidth]{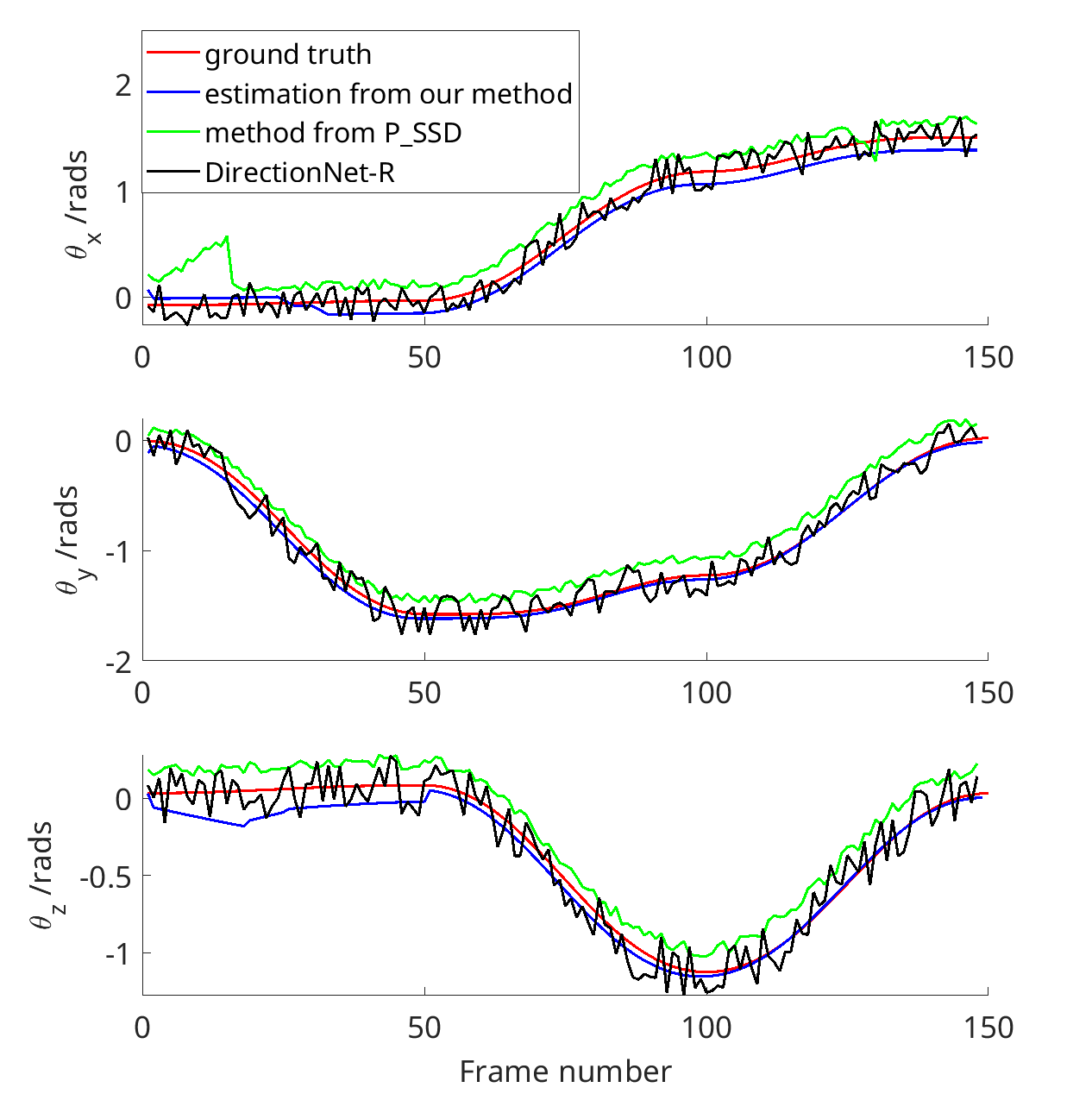}
    \caption{}
	\label{fig:seq_pure_3R} 
\end{subfigure}%
\begin{subfigure}{.33\textwidth}
  \centering
    \includegraphics[width=\linewidth]{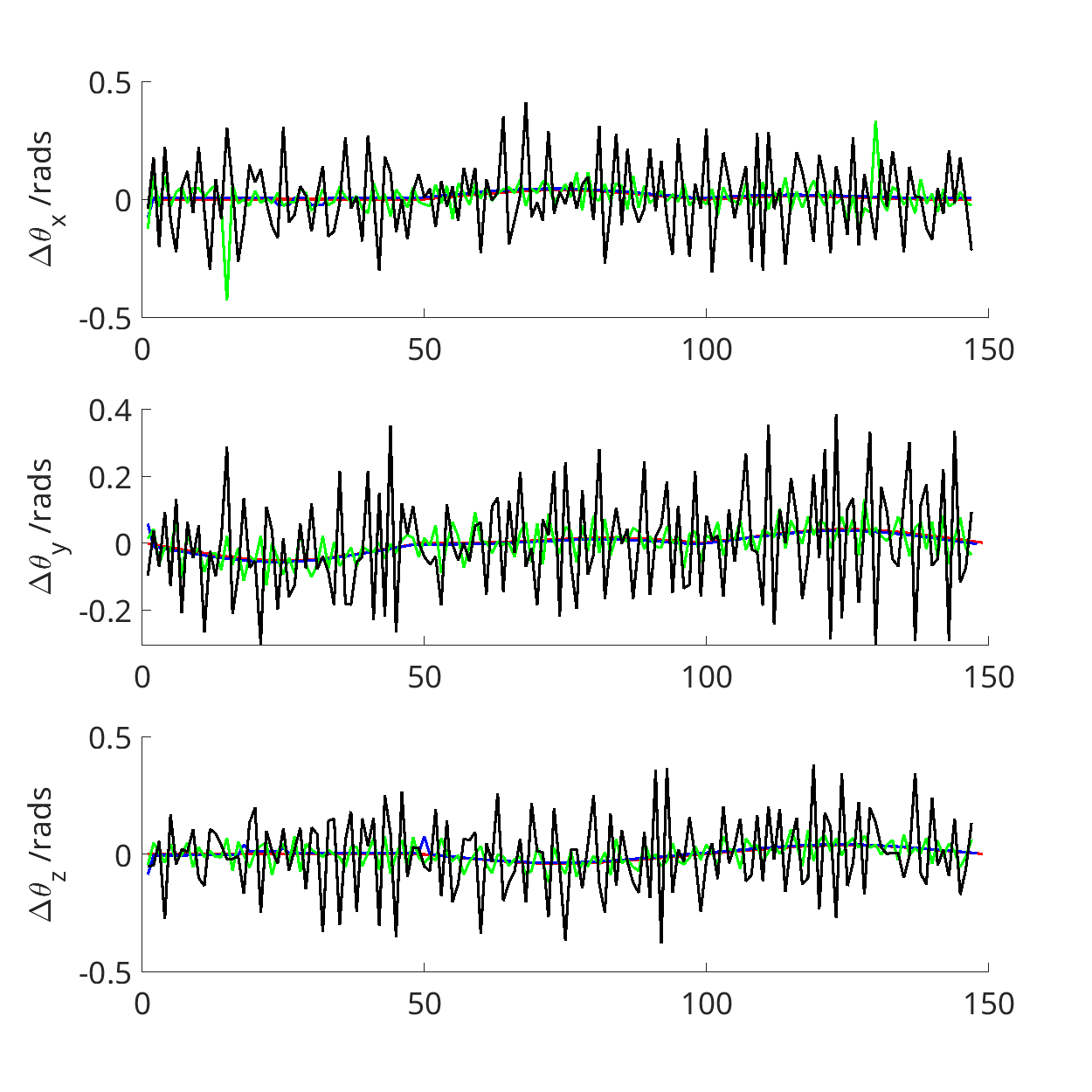}
    \caption{}
	\label{fig:seq_pure_3dR} 
\end{subfigure}%
\begin{subfigure}{.33\textwidth}
  \centering
    \includegraphics[width=\linewidth]{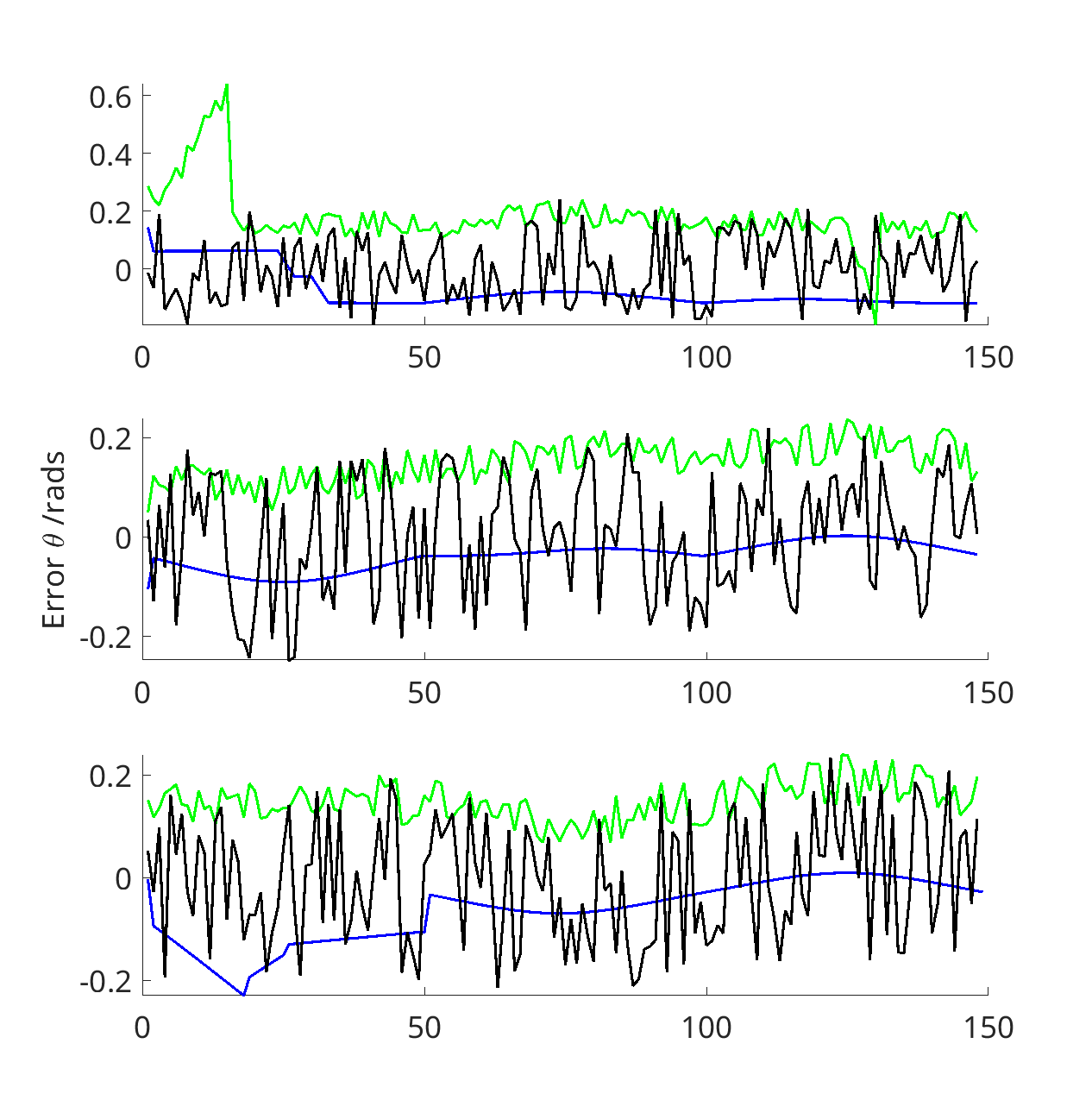}
    \caption{}
	\label{fig:seq_pure_err} 
\end{subfigure}%
\caption{{\bf Performance benchmark against ground-truth data and baseline VG approach in a simulated environment.} Our proposed analytical-LbTO VG method is evaluated against a baseline VG approach ~\cite{andre2022photometric,chen2021wide} to reveal higher accuracy and less noisy predictions. (a) Comparison of different approaches to estimate $\theta$, (b): Comparison of different approaches to estimate $\Delta \theta$, (c): The error of estimating the rotation angle. All comparisons are done along the three spatial axes for our method and the baseline VG method with ground-truth data shown for reference. The $x$, $y$, and $z$ rotations are expressed in terms of axis angles and plotted against the image number on the $x$-axis and the angle of rotation in radians on the $y$-axis. }
\label{fig:ResQuant}
\end{figure*}

\subsection{Comparison with Other Approaches}
We conducted a comparison between the method proposed in this paper and the methods presented in ~\cite{andre2022photometric,chen2021wide}. Where ~\cite{andre2022photometric} introduces a new visual gyroscope named P\_SDD using dual-fisheye cameras to accurately estimate orientation by projecting images onto a sphere, and ~\cite{chen2021wide} proposed DirectionNet, a novel approach to camera pose regression, estimates discrete distributions over a 5D relative pose space by factorizing camera pose into 3D direction vectors.

Figure \ref{fig:ResQuant} presents the results of comparison. Where Figure~\ref{fig:seq_pure_3R} depicts the estimated rotation angle in the world coordinate system for different frames in the sequence.
It can be observed that P\_SSD exhibits greater smoothness compared to DirectionNet. However, it has a relatively large cumulative error. Although DirectionNet does not possess a large cumulative error, it exhibits obvious Oscillation, indicating that the estimated value has a relatively large variance. Nevertheless, the method proposed in this paper has a smaller variance and higher accuracy. This is due to the utilization of a large number of masks. After undergoing MLP optimization, the method demonstrates stronger robustness.

The same conclusion can be drawn in Fig.~\ref{fig:seq_pure_3dR}, 
which shows the rotations between 2 images.
Fig.~\ref{fig:seq_pure_err} depicts the errors in estimation, which further confirm the effectiveness of the neural network model in accurately estimating the rotation. 

Our approach has been demonstrated to be more reliable in comparison to competing methods, as it exhibits a significantly higher mean error or larger variance. Besides, our comparative analysis of average errors shows that our method outperforms P\_SDD, with improvements in accuracy on the $xyz$ axes of [65\%, 10\%, 3\%], resulting in an overall accuracy increase of 26\%.

\section{Conclusion and Discussion}\label{discussion_5}
This paper highlights the advantages of using visual gyroscopes for 3D rotation estimation. Visual gyroscopes, being versatile and cost-effective, can easily integrate into devices like smartphones and autonomous robots without significant added cost or complexity.

The proposed approach combines the analytical solution of visual gyroscopes with machine learning optimization, demonstrating effectiveness and accuracy.
The machine learning optimization provides a crucial advantage, enhancing the accuracy.
Experiments show superior performance, surpassing traditional methods in accuracy and robustness when estimating rotation.

The results suggest that visual gyroscopes are a useful tool for applications in computer vision, robotics, and augmented reality. In computer vision, accurate 3D rotation estimation aids object tracking and image stabilization. In robotics, this precision is important for autonomous robots in dynamic environments. In augmented reality, precise 3D rotation estimation aligns virtual objects with the real world for a more immersive user experience.

In conclusion, our proposed approach for 3D rotation estimation using visual gyroscopes and machine learning optimization offers versatility, cost-effectiveness, accuracy, and robustness. Our contribution holds the potential to transform computer vision, robotics, and augmented reality and to make it a powerful and accessible technology for various applications.

\clearpage

\bibliographystyle{IEEEtran}
\bibliography{refs}

\end{document}